\documentclass[11pt]{article}
\usepackage{coling2020}
\usepackage{times}
\usepackage{url}
\usepackage{latexsym}

\usepackage{amsmath}
\usepackage{amssymb}
\usepackage{graphicx}

\usepackage{framed}
\usepackage{subfigure}
\usepackage{booktabs}
\usepackage{empheq}
\usepackage{array}
\usepackage{xcolor,soul}
\usepackage{enumitem}
\usepackage{multirow}
\usepackage{mathtools}

\usepackage{algorithm, algorithmicx, algpseudocode}

\DeclarePairedDelimiterX{\infdivx}[2]{(}{)}{%
	#1\;\delimsize\|\;#2%
}

\newcommand{\ourModelName}{\textsc{SSE}}

\newcommand{\cnndm}{\textit{CNN/Daily Mail}}
\newcommand{\rouge}{\textsc{Rouge}}

\colorlet{soulred}{red!50}
\colorlet{soulcyan}{cyan!50}

\DeclareRobustCommand{\hlcyan}[1]{{\sethlcolor{soulcyan}\hl{#1}}}
\DeclareRobustCommand{\hlred}[1]{{\sethlcolor{soulred}\hl{#1}}}

\newcommand{\citet}{\newcite}
\newcommand{\citep}{\cite}

\colingfinalcopy % Uncomment this line for the final submission

\title{At Which Level Should We Extract? An Empirical Analysis on Extractive Document Summarization}

\author{Qingyu Zhou\footnote{Contribution done during internship at Microsoft Research Asia.} \\
  Tencent Cloud Xiaowei \\
  Beijing, China \\
  {\tt qingyuzhou@tencent.com}  \\\And
  Furu Wei, Ming Zhou \\
  Microsoft Research Asia \\
  Beijing, China \\
  {\tt \{fuwei,mingzhou\}@microsoft.com} \\}

\date{}

\begin{document}
\maketitle
\begin{abstract}
Extractive methods have been proven effective in automatic document summarization.
Previous works perform this task by identifying informative contents at sentence level.
However, it is unclear whether performing extraction at sentence level is the best solution.
In this work, we show that unnecessity and redundancy issues exist when extracting full sentences, and extracting sub-sentential units is a promising alternative.
Specifically, we propose extracting sub-sentential units based on the constituency parsing tree.
A neural extractive model which leverages the sub-sentential information and extracts them is presented.
Extensive experiments and analyses show that extracting sub-sentential units performs competitively comparing to full sentence extraction under the evaluation of both automatic and human evaluations.
Hopefully, our work could provide some inspiration of the basic extraction units in extractive summarization for future research.
\end{abstract}

\blfootnote{
	%
	% for review submission
	%
	* Contribution done during internship at Microsoft Research Asia while pursuing PhD at Harbin Institute of Technology.
	\hspace{-0.65cm}  % space normally used by the marker
%	Place licence statement here for the camera-ready version. See
%	Section~\ref{licence} of the instructions for preparing a
%	manuscript.
	%
	% % final paper: en-uk version 
	%
	% \hspace{-0.65cm}  % space normally used by the marker
	% This work is licensed under a Creative Commons 
	% Attribution 4.0 International Licence.
	% Licence details:
	% \url{http://creativecommons.org/licenses/by/4.0/}.
	% 
	 % final paper: en-us version 
	
	 \hspace{-0.65cm}  % space normally used by the marker
	 This work is licensed under a Creative Commons 
	 Attribution 4.0 International License.
	 License details:
	 \url{http://creativecommons.org/licenses/by/4.0/}.
}

\section{Introduction}
Automatic text summarization aims to produce a brief piece of text which can preserve the most important information in it.
The important contents are identified and then extracted to form the output summary~\cite{nenkova2011automatic}.
In recent decades, extractive methods have proven effective in many systems~\cite{carbonell1998use,mihalcea2004textrank,mcdonald2007study,cao2015ranking,cheng-lapata:2016:P16-1,zhou-etal-2018-neural-document,nallapati2017summarunner}.

In previous works, extractive summarization systems perform extraction on the sentence level~\cite{mihalcea2004textrank,cheng-lapata:2016:P16-1,nallapati2017summarunner}.
As the extraction unit, a sentence is a grammatical unit of one or more words that express a statement, question, request, etc.
There are several advantages of extracting sentences.
First, extractive systems are simpler, easier to develop, and faster during run-time in real application scenarios, compared with abstractive systems.
Moreover, original sentences in the input are naturally fluent and grammatically correct.
Finally, extracted sentences are factually faithful to the input document, compared with abstractive methods~\cite{cao2018faithful}.

Despite the success of extractive systems, from previous works, it is still not clear whether extracting at sentence level is the best solution for extractive methods.
There are several drawbacks of extracting the full sentences.
The most obvious issue is that the extracted sentences may contain unnecessary information.
Some previous works have also noticed this problem and try to solve it by compressing or rewriting the extracted sentences~\cite{martins-smith-2009-summarization,chen-bansal-2018-fast,xu-durrett-2019-neural}.
Furthermore, extracted sentences may contain duplicate contents.
Thus, methods such as Maximal Marginal Relevance (MMR)~\cite{carbonell1998use} and sentence fusion~\cite{barzilay2005sentence,lebanoff-etal-2019-scoring} are proposed to avoid or merge duplicate contents.

The redundancy and unnecessity issues might be caused by extracting full sentences since an important sentence may also contain unnecessary information.
Besides, different importance sentences may have duplicate (un)important words.
This inspires us that we can perform extraction at a finer granularity so that the important and unimportant contents could be separated.
Therefore, we argue that extracting sub-sentential units in a sentence could be a solution.
As for the sub-sentential units, we mainly focus on the non-terminal nodes in a constituency parsing tree in this paper.
In a given parsing tree of a sentence, the root node represents the entire sentence while the leaf nodes represent each corresponding lexical token.
An extractive system could perform extraction on the non-terminal nodes which can express more fine-grained information.
To keep the advantages of extractive methods, we choose the nodes which can still express a full statement.
Specifically, the nodes with the clause tag such as S and SBAR are used for creating extraction units.

In this paper, we conduct experiments and analyses to answer the following questions:
\begin{enumerate}[label=Q\arabic*]
	\itemsep0em 
	\item Does extracting full sentences introduce unnecessary or duplicate information? (\S\ref{sec:q1})
	\item Can extracting sub-sentential units solve these problems of full sentence extraction? How to perform sub-sentential unit extraction? (\S\ref{sec:q2})
	\item Can extracting sub-sentential units improve the performance of extractive document summarization systems? (\S\ref{sec:experiment}, \S\ref{sec:autoeva}, \S\ref{sec:humaneva})
	\item Does extracting sub-sentential units cause any other issues? (\S\ref{sec:q4})
%	\item \underline{Is it safe to extract at an even finer granularity}.
\end{enumerate}
%Empirically, our main observations are summarized as:
%\begin{enumerate}[label=\arabic*)]
%	\item Performing extraction at the sentence-level is good solution, but it will introduce unnecessary and duplicate information
%	\item By extracting sub-sentential level units, 
%	\item Does extracting sub-sentential units improves the performance of summarization systems?
%	\item Does extracting sub-sentential units hurts the integrity of information.
%	\item \underline{Is it safe to extract at an even finer granularity}.
%\end{enumerate}

\section{Related Work}

Extractive document summarization methods have proven effective and been extensively studied for decades.
As an effective approach, extractive methods are popular and dominate the summarization research.
As the very first work in 1950s, \citet{luhn1958automatic} uses lexical frequency to determine the importance of a \textit{sentence}.
Many research works further develop extractive summarization methods.
The most important step is to determine the sentence importance.
Many different methods have been proposed and they can be roughly categorized from different perspectives.

From the perspective of having supervision or not, there are two major types: unsupervised methods and supervised methods.
One of the difficulties in training an extractive system is the lack of extraction labels.
The reason is that most of the reference summary is written by human experts, therefore, it is hard to find the exact appearance in the input document.
Without natural training labels, unsupervised and supervised methods treat extractive summarization as different problems.

%Unsupervised methods do not require model training or data annotation.
%In these methods, many surface features are useful, such as term frequency \cite{luhn1958automatic}, TF*IDF weights \cite{erkan2004lexrank}, sentence length \cite{cao2015ranking} and sentence positions \cite{Ren:2017:LCS:3077136.3080792}.
%These features can be used alone or combined with weights.

Graph-based methods \cite{erkan2004lexrank,mihalcea2004textrank,wan2006improved} are very useful unsupervised methods.
In these methods, the input document is represented as a connected graph.
The vertices represent the sentences, and the edges between vertices have attached weights that show the similarity of the two sentences.
The score of a sentence is the importance of its corresponding vertex, which can be computed using graph algorithms.

Supervised methods for extractive summarization create training labels manually.
\cite{cao2015ranking,Ren:2017:LCS:3077136.3080792} directly train regression models using \rouge{} scores as the supervision.
\cite{cheng-lapata:2016:P16-1,nallapati2017summarunner,zhou-etal-2018-neural-document,zhang-etal-2019-hibert} search the oracle extracted sentences as the training labels. 
\citet{cheng-lapata:2016:P16-1} propose  treating document summarization as a sequence labeling task.
They first encode the sentences in the document and then classify each sentence into two classes, i.e., extraction or not.
\citet{nallapati2017summarunner} propose a system called SummaRuNNer with more features, which also treat extractive document summarization as a sequence labeling task.
\citet{zhou-etal-2018-neural-document} propose using pointer networks~\cite{vinyals2015pointer} to repeatedly extract sentences.

Recently, Reinforcement Learning (RL) is also introduced in unsupervised extractive summarization~\cite{dong-etal-2018-banditsum,bohm-etal-2019-better}.
\citet{dong-etal-2018-banditsum} treat sentence extraction as a Bandit problem so that they can train a RL-based system whose reward is the \rouge{} scores.
\citet{bohm-etal-2019-better} propose that using human judgments as reward is better than \rouge{}.
However, these methods still need reward as a signal, which differ from the previously introduced fully unsupervised methods.

\section{Q1: Review of Extracting Full Sentence}
\label{sec:q1}
Performing sentence extraction in summarization systems have proven effective in previous works~\cite{luhn1958automatic,cao2015ranking,cheng-lapata:2016:P16-1,nallapati2017summarunner}.
Despite the success of these systems, it is still unclear whether performing content extraction at the sentence level is the best solution.
In this section, we will examine the  drawbacks of extraction at the sentence level. 

\subsection{The Dataset}
\begin{figure}[htbp]
	\centering
	\includegraphics[scale=0.6]{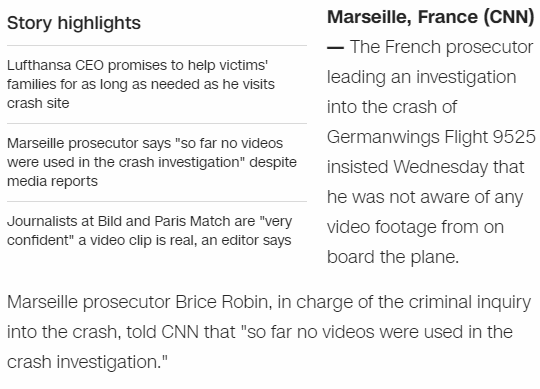}
	\caption{\label{fig:cnn} A screenshot example of the document-summary pair in the \cnndm{} dataset.}
\end{figure}

There are various datasets for text summarization, such as DUC/TAC, \cnndm{}, New York Times, etc.
In this paper, we take the most commonly used dataset in recent research works~\cite{cheng-lapata:2016:P16-1,nallapati2017summarunner,zhou-etal-2018-neural-document,xu-durrett-2019-neural,lebanoff-etal-2019-scoring}, \cnndm{}, as our testbed.
The statistics of it can be found in Table~\ref{table:data_stat}.
One of the most distinguishable features of this dataset is that the output summary is in the form of highlights written by the news editors.
As shown in the example in Figure~\ref{fig:cnn}, the summary (highlights) is a list of bullets.
Therefore, extractive methods perform well on this dataset~\cite{grusky2018newsroom}.

\subsection{The Drawbacks}
\label{sec:drawback}
There are two main potential drawbacks of extracting sentences.
First, unnecessary information is smuggled with the extracted sentences.
Second, duplicate content may appear when extracting multiple sentences.
%We mainly focus on these two problems in this section.
To analyze whether the issues exist, we conduct experiments and analyses with both count-based statistics and human judgments.
We consider two different settings to reach our final conclusion, i.e., the \textit{extractive oracle} and a real extractive system.

First, we check the quality of the \textit{sentence level extractive oracle}, since it is the upper bound of any extraction system.
Two different methods are used in recent extractive summarization research for building the oracle training label.
The first one is based on semantic correspondence~\cite{woodsend-lapata-2010-automatic} of document sentences and reference summary, used in \cite{cheng-lapata:2016:P16-1}.
The second one is heuristic, which maximizes the \rouge{} score with respect to gold summaries.
This one is more broadly used in many recent extractive systems~\cite{nallapati2017summarunner,zhou-etal-2018-neural-document,zhang-etal-2019-hibert,liu2019fine}.
We adopt the second method since it is more widely used and easy to implement.
The extractive oracle is computed with the metric of \rouge{}-2 F1 score, which is also the metric used in the final automatic evaluation in these systems.

Second, we check the output of a BERT-based sentence level extraction method and denote it as BERT-SENT.
Following previous works~\cite{devlin2018bert,liu2019fine,zhang-etal-2019-hibert}, BERT-SENT treats extractive document summarization as a sentence classification task.
The model is borrowed from \citet{liu2019fine}, but we remove the interval segment embeddings in it since it does not have obvious benefits.

%\red{Show one or two drawback examples here.}

\begin{table}[htbp]
%	\small
	\begin{center}
		\begin{tabular}{@{~}l@{\hspace{1ex}}c@{\hspace{1ex}}c@{\hspace{1ex}}c@{~}}
			\toprule
			& REF & Ora-sent & Ora-ss \\
			\midrule
			\# (Sent) & 3.88 & 2.61 & N/A \\
			\# (Word) & 58.31 & 70.46 &  52.77 \\
			\rouge{}-1 P & N/A & 52.59 & 61.84  \\
			\rouge{}-2 P & N/A & 33.97 &  43.45  \\
			%			\rouge{}-2 F1 & N/A & \\
			1-gram Overlap (\%)  & 15.77 & 19.24 & 16.75 \\
			2-gram Overlap (\%) &  1.40 & 2.22 & 1.90 \\
			3-gram Overlap (\%) & 0.21 & 0.51 & 0.45 \\
			\bottomrule
		\end{tabular}
	\end{center}
	\caption{\label{table:oracle_info}Statistics of the reference (REF), and the extractive oracle of sentence level (Ora-sent) and sub-sentential level (Ora-ss) on the \cnndm{} test set.}
\end{table}

\subsubsection{Unnecessary Information}
In this section, we examine whether unnecessary information is introduced unavoidably when extracting full sentences.
%We consider two different settings to reach our final conclusion.
%First, we check the quality of the \textit{oracle} training label, since it is the upper bound of any extraction system.
%Second, we check the output of a state-of-the-art extraction method, i.e., fine-tuning BERT~\cite{devlin2018bert,liu2019fine,zhang-etal-2019-hibert} as a sentence classification task.

\paragraph{Oracle:}
%Two different methods are used in recent extractive summarization research for building the oracle training label.
%The first one is based on semantic correspondence~\cite{woodsend-lapata-2010-automatic} of document sentences and reference summary, used in \cite{cheng-lapata:2016:P16-1}.
%The second one is heuristic, which maximize the \rouge{} score with respect to gold summaries.
%This one is more broadly used in many recent extractive systems~\cite{nallapati2017summarunner,zhou-etal-2018-neural-document,zhang-etal-2019-hibert,liu2019fine}.
%We adopt the second method since it is more widely used and easy to implement.
%The extractive oracle is computed with the metric of \rouge{}-2 F1 score, which is also the metric used in the final automatic evaluation in these systems.

Table~\ref{table:oracle_info} shows the information of the extractive oracle on the \cnndm{} test set.
The \rouge{}-1 precision of the extractive oracle is 52.59, which means that there are 47.41 percent of the unigrams are not in the reference summary.
As to the \rouge{}-2 scores, the precision drastically drops to 33.97.
These two metrics show that large amount of unwanted lexical units, i.e. unigram and bigram, are extracted along with the desired contents.
This indicates that there exists unnecessary information on the lexical level.

The surface lexical matching (\rouge{} scores) has its limits, that it cannot fully express the semantic level.
We also conduct human analysis to check whether unnecessary information is extracted at the same time.
The labeling criteria of unnecessary information is whether a 5-token span is not needed, lexically or semantically, comparing to ground-truth summary.
We randomly sampled 50 documents from the \cnndm{} test set.
Evaluation results show that, 48\% of the extractive oracles contain unnecessary information.

\paragraph{BERT-SENT:}
Similar experiments and analyses are also conducted on BERT-SENT.
Table~\ref{table:sse_info} shows the cound-based statistics.
Results show that 63.07\% percent of the unigrams and 82.73\% bigrams are not in the reference summary.
These rates are much higher than the sentence-level extractive oracle, and show that the unnecessity issue is quite severe.
Human evaluation shows that 54\% the outputs have unnecessary information.

\subsubsection{Redundancy}
In this section, we check whether redundancy problem exists in extractive summarization.
Similarly, we conduct the experiments and analysis on the extractive oracle and the BERT-SENT summarization system.
We first define a metric for redundancy, i.e., the n-gram overlap rate.
We calculate the n-gram overlap between each pair of sentences.
This overlap is calculated as:
\begin{equation}
\text{n-gram overlap} = 1 - \frac{\# (\text{unique n-gram})}{\# (\text{total n-gram})} 
\end{equation}

\paragraph{Oracle}
Table~\ref{table:oracle_info} shows the information of the extractive oracle on the \cnndm{} test set.
It can be observed that there are 19.24\% unigram and 2.22\% bigram are duplicated in the extractive oracle, which is much higher than the reference summmary.

Beyond this lexical level statistics, human evaluation is also conducted.
Results show that 12\% of the extractive oracle has the redundancy issue.
This result matches the n-gram overlap rates and shows that the redundancy issue even exists in oracle.

%The labeling criteria of redundancy is whether duplicate information exists in the summary.
%Evaluation results show that, \red{xx.xx} percent of the extractive oracle contain redundancy information.

\paragraph{BERT-SENT}
Results in Table~\ref{table:sse_info} shows that the BERT-SENT has high n-gram overlap rates, i.e., 27.18\% 1-gram and 7.68\% 2-gram overlap.
Thus, the redundancy issue is more severe in a real system than the extractive oracle, even for a state-of-the-art BERT-based system.
Human evaluation also shows that 49\% of the BERT-SENT output has the redundancy issue.

\section{Q2: Efficacy of Extracting Sub-sentential Units}
\label{sec:q2}
In this section, we propose an alternative to performing extraction on full sentence for extractive document summarization.
Instead, we perform extraction on sub-sentential units.
Specifically, the extraction units are based on the clause nodes in the constituency parsing tree of a sentence.
Figure~\ref{fig:tree} shows two simplified examples of the constituency tree.
The root node in a constituency tree represents the entire sentence, and the leaf node represents its corresponding lexical token.
Extracting on the root node is essentially extracting the full sentence, while extracting on the leaf node is doing compressing by extracting words~\cite{filippova2015sentence}.
We perform extraction on the non-terminal nodes which can both express a relatively complete meaning and be human-readable.
Therefore, the clause nodes, such as S and SBAR, become a good choice.
In this section, we introduce how to perform extraction on the sub-sentential units, and present a BERT-based model for it.

\subsection{The Sub-Sentential Units}
In order to perform extraction on the sub-sentential units, we need to determine what units can be extracted.
The proposed method is based on the constituency parsing tree.
The basic idea is based on the sub-sentential clauses in the tree.
In our experiments, we adopt the syntactic tagset used in the Penn Treebank (PTB)~\cite{Marcus:1993:BLA:972470.972475}.
There are two main types in the PTB tagset, phrase and clause.
We use  the clause tag since the information in a clause is more complete than a phrase.

\begin{figure}[H]
	\centering
	\includegraphics[scale=0.5]{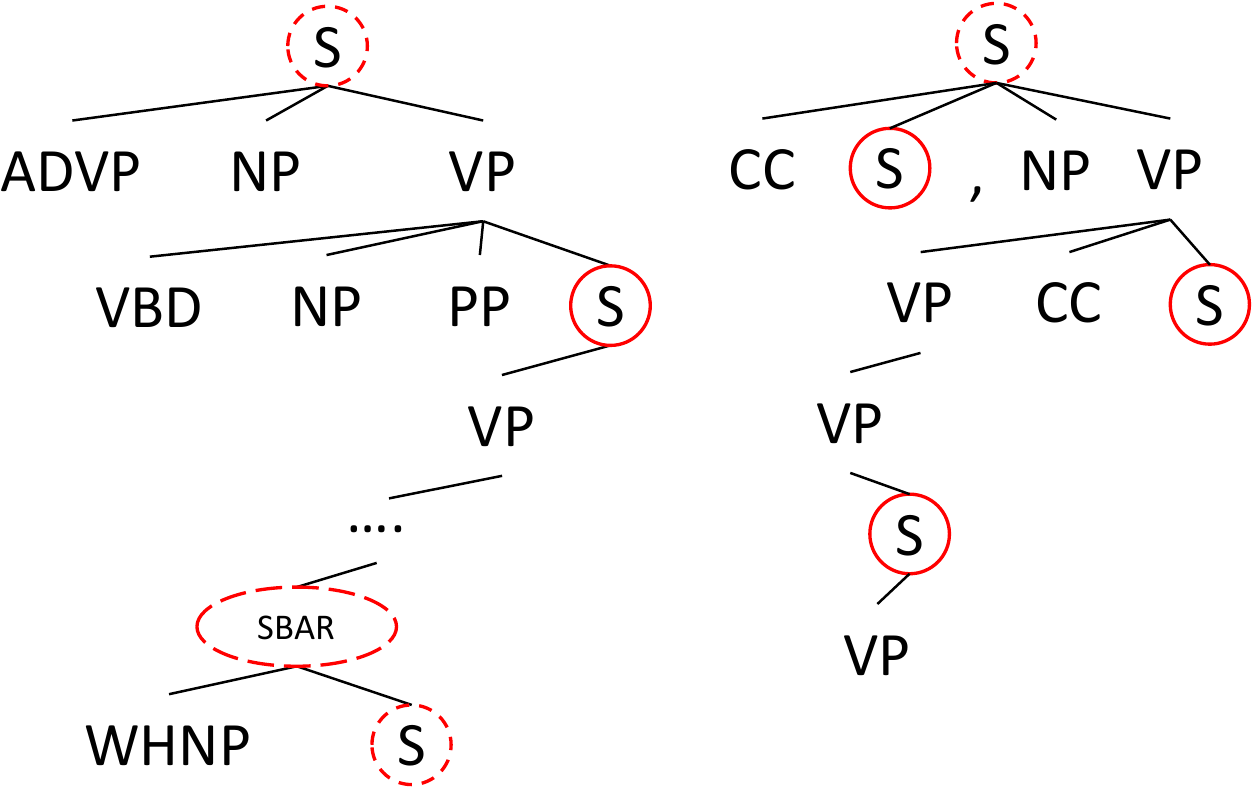}
	\caption{\label{fig:tree} Two simplified constituency parsing trees.
		The nodes in circles are candidates.
		The final selected node is the on in red solid-lined circle.
	}
\end{figure}

Given the parsing tree $ t_{i} $ of sentence $ s_{i} $, we traverse it to determine the boundary of extraction units.
%Figure~\red{x} shows an example of this procedure.
Specifically, every clause is treated as the extraction unit candidates.
If one of its ancestors is a clause node, we choose the highest level ancestor clause node  (except for the root node) as the extraction unit to include more complete information.
This heuristic is visualized in Figure~\ref{fig:tree}.
If no sub-sentential clauses can be found in a sentence, we use the full sentence as the extraction unit.
Finally, the input sentence is split into chunks using the selected clauses' boundaries.

%\begin{algorithm}[!htb]
%	\caption{Sub-Sentential Units Extraction}
%	\label{alg:datacreation}
%	%\hspace*{\algorithmicindent} 
%	\textbf{Input:} The constituency parsing tree $ t $, ; 
%	% \hspace*{\algorithmicindent} \textbf{Output:} The dialogue policy ; \\
%	\begin{algorithmic}[1]
%		\Procedure{training process}{}
%		\While{$k < m$}
%		\State $b_k \leftarrow$ Scheduler($b$, $m$, $k$);
%		\Repeat
%		\State $g^{u} \leftarrow$ UserGoalSampler($A$);
%		\State $B^{r}, B^{s}, c^{u} \leftarrow$ Controller($g^{u}$, $b_k$, $A$, $W$);
%		\State $b_k \leftarrow b_k - c^{u}$;
%		\Until {$b_k \leq 0 $}
%		\State Train the dialogue agent $A$ on $B^{r} \cup B^{s}$
%		\State Train world model $W$ on $B^{r}$
%		\EndWhile
%		\EndProcedure
%	\end{algorithmic}
%\end{algorithm}

\begin{figure}[htbp]
	\centering
	\includegraphics[scale=0.35]{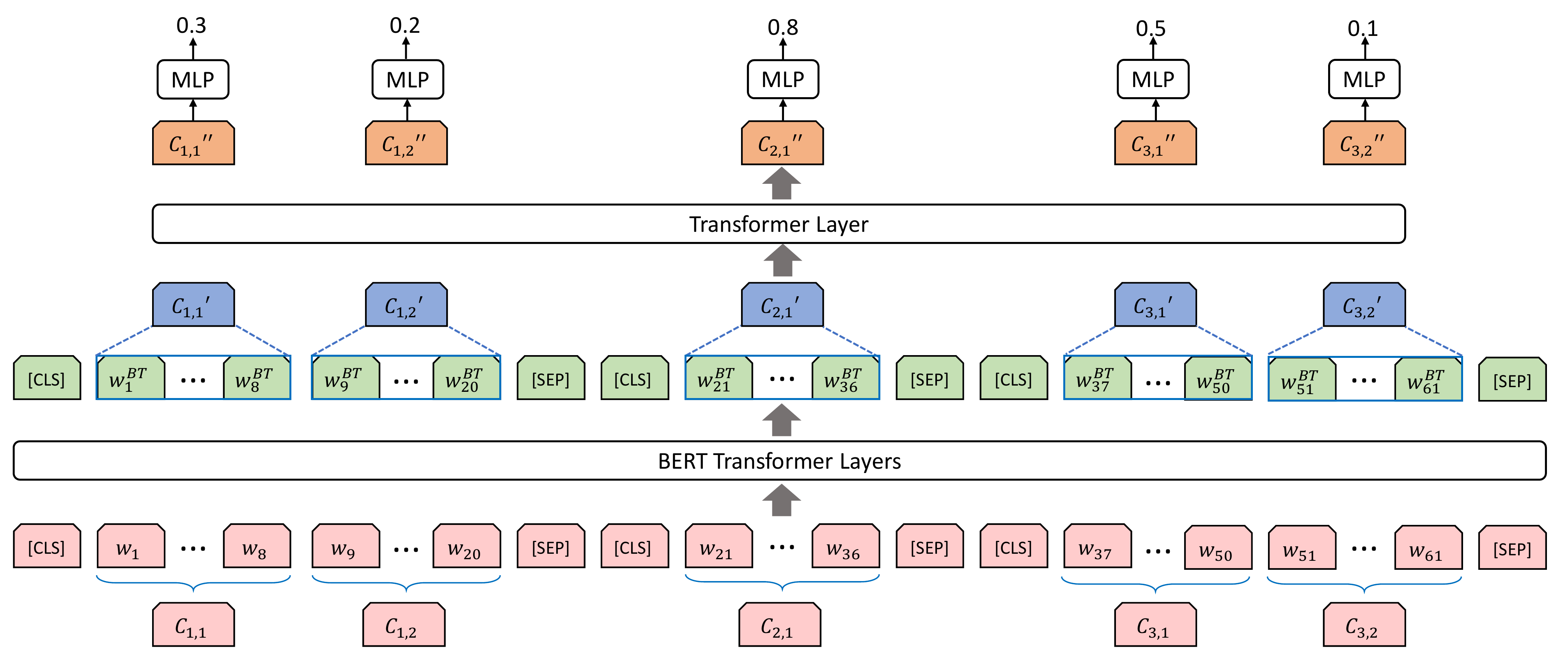}
	\caption{\label{fig:model} The overview of the BERT-based model for \textbf{s}ub-\textbf{s}entential \textbf{e}xtraction (SSE).
		In this simplified example, the document has 3 sentences. The first and the third sentences have two extraction units and the second sentence has one.
		After encoding the document with pre-trained BERT encoder, an average pooling layer are used to aggregate information of each extraction unit.
		The final Transformer layer captures the document-level information and then the MLP predicts the extraction probability.
	}
\end{figure}

\subsection{Model}
In this section, we present a BERT-based neural extractive summarization model for extracting \textbf{s}ub-\textbf{s}entential \textbf{u}nits (SSE).
We follow previous works~\cite{cheng-lapata:2016:P16-1,nallapati2017summarunner,xu-durrett-2019-neural,liu2019fine} to treat the document summarization as a sequence labeling task.
Figure~\ref{fig:model} shows the overview of the proposed model.
It consists of two levels of encoders.
The first level is the BERT-based document encoder, and the second level is the Transformer-based sub-sentential units encoder.
The BERT-based document encoder reads the tokens in the document, and then the Transformer-based encoder constructs the final extraction unit representations.

\subsubsection{BERT-based Encoder}
Following previous works~\cite{liu2019fine,zhang-etal-2019-hibert,lebanoff-etal-2019-scoring} which use BERT and achieve state-of-the-art results, we use BERT as the first level encoder.
The processed input document is denoted as $ \mathcal{D} = (S_{1}, S_{2}, \dots, S_{n}) = (w_{1}, w_{2}, \dots, w_{m})$ with $ n $ sentences, $ m $ BPE tokens.
The $ i $-th sentence contains $ l $ chunks $ S_{i} = (C_{i,1}, C_{i,2}, \dots, C_{i,l}) $.
The $ j $-th chunk with $ k $ words in $ S_{i} $ is denoted as $ C_{i,j} = (w_{i, j,1}, \dots, w_{i,j,k}) $.
Following \citet{liu2019fine}, we add additional \texttt{[CLS]} and \texttt{[SEP]} labels between sentences to separate them.
However, since the extraction unit is not the full sentence, the vector of  \texttt{[CLS]} is not used for classification in our model.
After the BERT encoder, the vector of the $ m $ document tokens are represented as $ (w_{1}^{BT}, w_{2}^{BT}, \dots, w_{m}^{BT}) $.

\subsubsection{Transformer-based Encoder}
The BERT-based encoder reads the entire document and builds the representation of each words.
The Transformer-based encoder then constructs the final representation of each chunk.
As shown in Figure~\ref{fig:model}, we first apply an average pooling on the chunk level.
Specifically, given the BERT-based encoder output of chunk $ C_{i,j} = (w_{i, j,1}^{BT}, \dots, w_{i,j,k}^{BT}) $, the pooled representation $ C_{i,j}' $ is:
\begin{equation}
C_{i,j}' = \frac{1}{k} \sum_{1}^{k} w_{i,j,}
\end{equation}
To note that, the \texttt{[CLS]} and \texttt{[SEP]} labels are not covered by the chunks, and thus not used in the average pooling.

After the average pooling, the document is represented as a sequence of chunk vectors: 
$ \mathcal{C'} = (C_{1,1}', \dots, C_{1,l1}', \dots, C_{n,1}', \dots, C_{n,ln}') $.
We then apply a chunk level Transformer to capture their relationship for extracting summaries:
\begin{empheq}{align}
	\mathcal{\hat{C}'} &= \text{LN}\left( \mathcal{C'} + \text{MultiHead}(\mathcal{C'}) \right)\\
	\mathcal{C''} &= \text{LN}\left(\mathcal{\hat{C}'} + \text{FFN}(\mathcal{\hat{C}'})\right)
\end{empheq}
where $ \text{MultiHead}(\cdot) $ is Multi Head Attention in Transformer~\cite{vaswani2017attention}, $ \text{LN}(\cdot) $ is Layer Normalization~\cite{ba2016layer}, $ \text{FFN}(\cdot) $ is a feed-forward network which consists of two linear transformations with a ReLU activation in between.
In this paper, we simplify the $ \text{MultiHead}(Q,K,V) $ to $ \text{MultiHead}(\cdot) $ since we only use the self attention mechanism for encoding thus $ Q=K=V $.

\subsubsection{Training Objective}
With the chunk level representation vectors $ \mathcal{C''} $, the model predict the output probability of each chunk $ C_{i,j}'' $:
\begin{equation}
p(C_{i,j}'') = \sigma(\mathbf{W}_{o}C_{i,j}'' + b_{o})
\end{equation}
where $ \sigma(\cdot) $ is the sigmoid function.
%, $  \mathbf{W}_{o} $ and $ b_{o} $ are weight parameters of a linear layer.
The training objective of the model is the binary cross-entropy loss given the extractive oracle label $ y_{i,j} $ and the predicted probability $ p(C_{i,j}'') $.
%\begin{empheq}{align}
%\mathcal{L} = - &\sum_{i=1}^{n} \left( \sum_{j=1}^{l} \left( y_{i,j}\cdot \log p(C_{i,j}'') \right. \right. \\
%&+ \left. \left. (1-y_{i,j})\cdot \log (1 - p(C_{i,j}''))  \right) \vphantom{\int_1^2} \right)
%\end{empheq}
%\begin{equation}
%\begin{split}
%\mathcal{L} = - \sum_{i=1}^{n} \left( \sum_{j=1}^{l} \left( y_{i,j}\cdot \log p(C_{i,j}'') \right. \right. 
%+ \left. \left. (1-y_{i,j})\cdot \log (1 - p(C_{i,j}''))  \right) \vphantom{\int_1^2} \right)
%\end{split}
%\end{equation}

\section{Experiment}
\label{sec:experiment}

\subsection{Dataset}
Following previous extractive  works~\cite{zhou-etal-2018-neural-document,xu-durrett-2019-neural,lebanoff-etal-2019-scoring,zhang-etal-2019-hibert,dong-etal-2018-banditsum}, 
we conduct data preprocessing using the same method\footnote{https://github.com/abisee/cnn-dailymail} in \citet{see-liu-manning:2017:Long}, including sentence splitting and word tokenization.
we preprocess the data as same as \citet{see-liu-manning:2017:Long}.
We then use a state-of-the-art BERT-based constituency parser \cite{Kitaev-2018-SelfAttentive} to process the input document whose performance is 95.17 F1 on WSJ test set.
The statistic of the original \cnndm{} dataset and the sub-sentential version are listed in Table~\ref{table:data_stat}.

\begin{table}[htbp]
%	\small
	\begin{center}
		\begin{tabular}{@{~}l@{\hspace{1ex}}c@{\hspace{1ex}}c@{\hspace{1ex}}c@{~}}
			\toprule
			\bf \cnndm{}       & \bf Training & \bf Dev & \bf Test \\ 
			\midrule
			\#(Document)       & 287,227      & 13,368  &  11,490 \\
			\#(Ref / Document) & 1            & 1       & 1 \\
			Doc Len (Sentence) & 31.58        & 26.72   & 27.05 \\
			Doc Len (Word)     & 791.36       & 769.26  & 778.24 \\
			Ref Len (Sentence) & 3.79         & 4.11    & 3.88   \\
			Ref Len (Word)     & 55.17        & 61.43   & 58.31 \\
			\midrule
			\small{Doc Len (Sub-Sentence)} &    52.84     &  51.37   &  52.02 \\
			\bottomrule
		\end{tabular}
	\end{center}
	\caption{\label{table:data_stat}Data statistics of \cnndm{} dataset.}
\end{table}

\subsection{Implementation Details}
We found that the tokenizer used in the constituency parser is different from the one in BERT.
Therefore, we apply some simple tokenization fix to process the text before feeding them into BERT.
The input of BERT-based encoder is then processed with the BERT's subword tokenizer.
Since the maximum length in the BERT's position embedding is 512, we truncated the document to 512 subwords.
We use Adam \citep{kingma2014adam} as the optimizing algorithm.
For the hyperparameters of Adam optimizer, we set the learning rate $ \alpha = 2e-5 $, two momentum parameters $ \beta_{1} = 0.9 $ and $ \beta_{2} = 0.999 $ respectively, and $ \epsilon=10^{-8} $.
The model is implemented with PyTorch \citep{paszke2017automatic} and PyTorch Transformer~\cite{Wolf2019HuggingFacesTS}.
We use the  \texttt{bert-base-uncased} version of BERT.
%, which has 12 pre-trained Transformer layers.
We train the model using 4 NVIDIA P100 GPUs with a batch size of 40.
The dropout~\cite{srivastava2014dropout} rates in all the Transformer layers are set to 0.1.
We train the model for 4 epochs which takes about 6 hours.
The final model is picked according to the performance on the development set  among the 4 model checkpoints.

During inference, we rank the extraction units according to $ p(C_{i,j}'') $ and select the top ones.
Since the extraction unit in this paper is shorter than full sentence, we repeatedly select next sub-sentential unit until the summary length reaches the limit.
The length limit is set to 60 words according to the statistics on the development set in Table~\ref{table:data_stat}.

\subsection{Evaluation Metric}
We employ \textsc{Rouge} \citep{lin2004rouge} as our evaluation metric.
%\textsc{Rouge} measures the quality of summary by computing overlapping lexical units, such as unigram, bigram, trigram, and longest common subsequence (LCS).
%It has become the standard evaluation metric for DUC shared tasks and popular for summarization evaluation.
Following previous work, we use \textsc{Rouge}-1 (unigram), \textsc{Rouge}-2 (bigram) and \textsc{Rouge}-L (LCS) as the evaluation metrics in the experimental results.

Additionally, we also conduct human evaluation on the output summaries.
Following previous works~\cite{cheng-lapata:2016:P16-1,nallapati2017summarunner,liu2019fine,zhang-etal-2019-hibert}, we randomly sampled 50 documents from the \cnndm{} test set, which is the same as in \S\ref{sec:q1}.

\section{Results}

\subsection{Automatic Evaluation}
\label{sec:autoeva}
Table~\ref{table:cnndm_result} shows the \rouge{} evaluation results.
We compare the \ourModelName{} with the following systems:

\paragraph{Abstractive Systems} 
%Pointer-Generator Network (PGN)~\cite{see-liu-manning:2017:Long} and DCA~\cite{celikyilmaz-etal-2018-deep} are sequence-to-sequence models with copy and coverage mechanisms.
Pointer-Generator Network (PGN)~\cite{see-liu-manning:2017:Long} ia a sequence-to-sequence model with copy and coverage mechanisms.
FastRewrite~\cite{chen-bansal-2018-fast} conducts extraction first then generation.
JECS~\cite{xu-durrett-2019-neural} first extracts sentences then compresses them to reduce redundancy.
%InconsisLoss~\cite{hsu-etal-2018-unified} regularizes the word level attention with sentence level extraction attention.
Bottom-Up~\cite{gehrmann-etal-2018-bottom} applies constrains on the copying probability.
\paragraph{Extractive Systems}
\textsc{LEAD3} is a commonly used baseline which simply extracts the first three sentences.
\textsc{TextRank}~\cite{mihalcea2004textrank} is a popular graph-based unsupervised system.
\textsc{SummaRuNNer} and \textsc{NN-SE}~\cite{nallapati2017summarunner,cheng-lapata:2016:P16-1} use hierarchical structure for document encoding and predict sentence extraction probabilities.
%\textsc{LatentSum}~\cite{zhang-etal-2018-neural-latent},\textsc{Refresh}~\cite{narayan-etal-2018-ranking} and \textsc{BanditSum}~\cite{dong-etal-2018-banditsum} leverage reinforcement learning in extractive summarization.
\textsc{NeuSum}~\cite{zhou-etal-2018-neural-document} jointly model the sentence scoring and selection steps.
\textsc{BertSumExt}, \textsc{BertSumExt + TriBlk} (with trigram blocking)~\cite{liu2019fine}, \textsc{Self-Supervised}~\cite{wang2019selftrain} and \textsc{HIBERT}~\cite{zhang-etal-2019-hibert} use pre-training techniques in extractive document summarization.
BERT-SENT is the sentence-level extractive baseline described in section~\ref{sec:drawback}~.

\begin{table}[h]
%		\small
	\begin{center}
		%		\begin{tabular}{lccc}
		\begin{tabular}{@{~}l@{\hspace{1ex}}c@{\hspace{1ex}}c@{\hspace{1ex}}c@{~}}
			\toprule
			\bf Model & \bf {\small \textsc{Rouge-1}} & {\small \bf \textsc{Rouge-2}} & {\small \bf \textsc{Rouge-L}} \\ 
			\hline
			%			\small{\textit{Abstractive}} & & & \\
			\hline
			\textsc{PGN}  & 39.53  &  17.28 &  36.38 \\
%			\textsc{DCA} & 41.69 & 19.47 & 37.92 \\
			FastRewrite  & 40.88 & 17.80 & 38.54 \\
			JECS & 41.70 &18.50 &37.90\\
%			InconsLoss  & 40.68 & 17.97 & 37.13 \\
			Bottom-Up & 41.22 & 18.68 & 38.34\\
			\hline
			%			\small{\textit{Extractive}} & & & \\
			\hline
			\textsc{LEAD3}  & 40.24  & 17.70 & 36.45  \\
			\textsc{TextRank} &  40.20 & 17.56  & 36.44 \\
			{\small \textsc{SummaRuNNer}}  & 39.60 &  16.20  & 35.30 \\
			%			\textsc{CRSum}  & 40.52\significant{} & 18.08\significant{}  & 36.81\significant{} \\
			\textsc{NN-SE}    & 41.13  &  18.59  & 37.40   \\
			\textsc{NeuSum} & 41.59 & 19.01 & 37.98  \\
%			\textsc{LatentSum} & 41.05 & 18.77  & 37.54\\
%			\textsc{Refresh}  & 40.00  &  18.20  & 36.60   \\
%			\textsc{BanditSum} & 41.50 &18.70 &37.60\\
			
			\textsc{BertSumExt}  & 42.61  &  19.99 &  39.09 \\
			\textsc{BertSumExt+TriBlk} & \textbf{43.25} &  20.24  &  39.63  \\
			\textsc{Self-Supervised}& 41.36 &   19.20  &  37.86  \\
			\textsc{HIBERT}  &  42.10 &  19.70  &  38.53 \\
%			\hline
			%			\small{\textit{Extractive}} & & & \\
			\hline
			\hline
			BERT-SENT & 42.13 & 19.73 & 38.59 \\
			\ourModelName{}  & 42.72  &  \textbf{20.29}   &  \textbf{39.98}  \\
			\bottomrule
		\end{tabular}
	\end{center}
	\caption{\label{table:cnndm_result}Full length \textsc{Rouge} F1  evaluation (\%) on CNN / Daily Mail test set. 
%		\rouge{}-1, 2 and L are reported.
%		Results with \otherpaper{} mark are taken from the corresponding papers. 
		%		Those marked with * were trained and evaluated on the anonymized dataset, and so are not strictly comparable to our results on the original text. 
%		All our \textsc{Rouge} scores have a 95\% confidence interval of at most $ \pm $0.22 as reported by the official ROUGE script.
%		The improvement is statistically significant with respect to the results with superscript \significant{} mark.	 
 Results for comparison systems are taken from the authors’ respective papers or obtained 
% on our data 
 by running publicly released software.
	}
\end{table}

The proposed method \ourModelName{} achieves the state-of-the-art results on the \cnndm{} dataset.
According to the output of the official \rouge{} script, the difference between \ourModelName{} and baselines are all statistically significant with a 0.95 confidence interval.
Compared to our sentence-level extraction baseline system BERT-SENT, using sub-sentential unit extraction leads to a +0.56 \rouge{} improvement.
As for the other existing systems which leverage BERT or other pre-training techniques and perform extraction on sentence level, \ourModelName{} still outperforms them statistically significantly in terms of \rouge{}.

\subsection{Human Evaluation}
\label{sec:humaneva}

Human evaluations are also conducted on the same 50 randomly sampled documents as in Section~\ref{sec:q1}.
The BERT-SENT and \ourModelName{} models are evaluated.
The workers are asked to rank the outputs of these systems from best to worst by the overall quality (with ties allowed).
In addition, we are also curious about how sub-sentential extraction solves the problems of full sentence extraction.
Specifically, the workers are asked to identify whether redundant or unnecessary information exists.

\begin{table}[htbp]
%	\small
	\begin{center}
		\begin{tabular}{l|cc}
			\toprule
			& BERT-SENT & \ourModelName{} \\ 
			\midrule
			Unnecessity  & 54\% & 37\% \\
			Redundancy & 49\% & 29\% \\
			\midrule
			Readability  & 1.00  & 1.24 \\
			Overall  &  1.50  & 1.35 \\
			\bottomrule
		\end{tabular}
	\end{center}
	\caption{\label{table:humaneva}Human evaluation results.
		Unnecessity and Redundancy are reported as occurrence frequency, and lower is better.
		Readability and Overall are reported as ranking, and lower is better.
	}
\end{table}

\begin{table*}[htb]
		\small
	\begin{center}
		%		\begin{tabular}{lccc}
		\begin{tabular}{p{0.98\textwidth}}
		\toprule
%		\hline
		\textit{\textbf{Document:}} (CNN) When \hlcyan{\textit{Etan Patz went missing in New York City at age 6}}, hardly anyone in America could help but see his face at their breakfast table. His photo's appearance on milk cartons after his May 1979 disappearance marked an era of heightened awareness of crimes against children. On Friday, more than 35 years after frenzied media coverage of his case horrified parents everywhere, \hlcyan{\textit{a New York jury will again deliberate over a possible verdict against the man charged in his killing, Pedro Hernandez.}} \hlred{He confessed to police three years ago}. \hlcyan{\textit{Etan Patz's parents have waited that long for justice}}, but some have questioned whether that is at all possible in Hernandez's case. ...... He said he killed the boy and threw his body away in a plastic bag. Neither the child nor his remains have ever been recovered. \hlcyan{\textit{But Hernandez has been repeatedly diagnosed with schizophrenia and has an ``IQ in the borderline-to-mild mental retardation range, ''}} his attorney Harvey Fishbein has said. Police interrogated Hernandez for 7 1/2 hours before he confessed. ...... A judge found \hlcyan{\textit{Ramos responsible for the boy's death}} and ordered him to pay the family \$ 2 million -- money the Patz family has never received. ...... \\
%		\hline
		\midrule
		\textit{\textbf{Reference}:}  The young boy 's face appeared on milk cartons all across the United States . Patz 's case marked a time of heightened awareness of crimes against children . Pedro Hernandez confessed three years ago to the 1979 killing in . \\
		\bottomrule
%		\hline
		\end{tabular}
	\end{center}
	\caption{\label{table:casestudy}An example document and gold summary in the \cnndm{} test set.
		The \hlred{words highlighted with red}  are extracted as a full sentence.
		The \hlcyan{\textit{italic words highlighted with cyan}} are extracted as sub-sentential units.
	}
\end{table*}

Table~\ref{table:humaneva} presents the human evaluation results.
We compare the \ourModelName{} with BERT-SENT.
As shown in the results, \ourModelName{} performs better than BERT-SENT for both redundancy and unnecessity.
The frequency of having these issues drops 20\% and 17\% respectively.
Thus the overall quality of \ourModelName{} is also better than BERT-SENT.

\subsection{Analysis}
\label{sec:analysis}
Table~\ref{table:casestudy} shows an example of a document, gold summary and the output of \ourModelName{}.
It can be observed that by performing sub-sentential extraction, the full sentence is broken into more fine-grained semantic units.
Therefore, the model can extract important parts without introducing unimportant contents.

Table~\ref{table:sse_info} shows the statistics of the outputs of BERT-SENT and \ourModelName{}.
Similarly, we conduct experiments and analyses with both statistics and human judgments, on both the unnecessary information and redundancy issues as in Section~\ref{sec:q1}.
Compared to BERT-SENT, \ourModelName{} performs significantly better in terms of \rouge{} precision by a large margin.
This shows that extracting sub-sentential units can bring less unimportant information.
We also found that the n-gram overlap rate of \ourModelName{} is also much lower than BERT-SENT, which shows that the output contains less redundant contents.

\begin{table}[htbp]
%	\small
	\begin{center}
		\begin{tabular}{lcc}
			\toprule
			& BERT-SENT & \ourModelName{} \\
			\midrule
			\# (Sent) & 3 & N/A \\
			\# (Word) & 82.83 & 68.74 \\
			\rouge{}-1 P & 36.93  &  39.22 \\
			\rouge{}-2 P & 17.27 & 18.66 \\
			1-gram Overlap (\%)  & 27.18 & 24.58 \\
			2-gram Overlap (\%) & 7.68 & 6.00 \\
			3-gram Overlap (\%) & 4.13 & 2.85 \\
			\bottomrule
		\end{tabular}
	\end{center}
	\caption{\label{table:sse_info}Statistics of the BERT-SENT and \ourModelName{} methods on the \cnndm{} test set.}
\end{table}

%Same human evaluation as in Section~\red{x} is also conducted on the output summary of \ourModelName{}.

\subsection{Q4: Readability of Sub-Sentential Units}
\label{sec:q4}
Performing sub-sentential unit extraction improves the \rouge{} scores and alleviates the problem of extracting full sentences.
However, it is not clear whether this method introduces new issues.
One possible issue is that the sub-sentential units are fragmented so the readability is poor.
To investigate this problem, we also add an item about the readability of the produced summary in the human evaluation questionnaire.
In detail, the workers are asked to rank the system outputs by the readability (with ties allowed).

Table~\ref{table:humaneva} shows the results of readability.
As shown in results, the BERT-SENT is always ranked as the best since the sentences are fully extracted.
The readability of extracted sub-sentential units is slightly worse than the full sentences.
We also manually checked the output of \ourModelName{} whose readability is worse.
We found that there are two reasons: 1) the sub-sentence is fragmented which affects the readability; 2) the sub-sentence is wrongly extracted due to the error of the constituency parser.
Therefore, we hope that the readability of \ourModelName{} could be improved if we can: 1) design better sub-sentential unit extraction algorithm; 2) have an even better syntactic parser.

\section{Conclusion}
In this paper, we investigate the problem of the extraction granularity for extractive document summarization.
We observe that performing extraction at sentence level has the redundancy and unnecessity issues.
We found that these problems can be alleviated by doing sub-sentential unit extraction.
Both automatic and human evaluations show that sub-sentential extraction performs competitively compared to the full-sentence-extraction systems.
Therefore, sub-sentential unit extraction could be a promising alternative to full-sentence extraction.
Our experiments and analyses on revisiting the basic extraction unit could provide some hints for future research on this direction.

%\section*{Acknowledgements}
%
%We thank the anonymous reviewers for their helpful
%comments.

\bibliographystyle{coling}
\bibliography{coling2020}

\end{document}